\documentclass[
]{ceurart}

\usepackage{listings}
\usepackage{framed}
\usepackage{multirow}

\let\OldFrameCommand\FrameCommand

\renewcommand{\FrameCommand}{%
  {\color{black}\vrule width 0.4pt\hspace{0.5em}}%
  \fboxsep=0pt
}

\usepackage{subfigure}

\newcommand{\tabindent}{\hspace*{10pt}}

\begin{document}

\copyrightyear{2025}
\copyrightclause{Copyright for this paper by its authors.
  Use permitted under Creative Commons License Attribution 4.0
  International (CC BY 4.0).}

\conference{IntRS'25: Joint Workshop on Interfaces and Human Decision Making for Recommender Systems, September 22, 2025, Prague}

\title{From latent factors to language: a user study on LLM-generated explanations for an inherently interpretable matrix-based recommender system}

\author[1]{Maxime Manderlier}[%
email=maxime.manderlier@umons.ac.be,
orcid=0000-0002-5641-9818]
\cormark[1]

\author[1]{Fabian Lecron}[%
email=fabian.lecron@umons.ac.be,
orcid=0000-0002-6516-0086]

\author[2]{Olivier Vu~Thanh}[%
email=olivier.vuthanh@umons.ac.be,
orcid=0000-0002-4917-6571]

\author[2]{Nicolas Gillis}[%
email=nicolas.gillis@umons.ac.be,
orcid=0000-0001-6423-6897]

\address[1]{Department of Technological Innovation Management, Faculty of Engineering, University of Mons (UMONS), Mons, Belgium}
\address[2]{Department of Mathematics and Operational Research, Faculty of Engineering, University of Mons (UMONS), Mons, Belgium}

\cortext[1]{Corresponding author.}

\begin{abstract}
We investigate whether large language models (LLMs) can generate effective, user-facing explanations from a mathematically interpretable recommendation model. 
The model is based on constrained matrix factorization, where user types are explicitly represented and predicted item scores share the same scale as observed ratings, making the model’s internal representations and predicted scores directly interpretable. This structure is translated into natural language explanations using carefully designed LLM prompts.
Many works in explainable AI rely on automatic evaluation metrics, which often fail to capture users’ actual needs and perceptions. 
In contrast, we adopt a user-centered approach: we conduct a study with 326 participants who assessed the quality of the explanations across five key dimensions---transparency, effectiveness, persuasion, trust, and satisfaction---as well as the recommendations themselves.
To evaluate how different explanation strategies are perceived, we generate multiple explanation types from the same underlying model, varying the input information provided to the LLM. 
Our analysis reveals that all explanation types are generally well received, with moderate statistical differences between strategies. User comments further underscore how participants react to each type of explanation, offering complementary insights beyond the quantitative results.

\end{abstract}


\maketitle

\section{Introduction}

Recommender systems have become essential tools for helping users navigate large catalogs of content. However, the algorithms powering these systems are often opaque, making it difficult for users to understand or trust their suggestions. This has motivated a growing body of work in explainable recommendation, which seeks to provide users with reasons for each recommendation. While many existing approaches generate explanations post hoc—often disconnected from the underlying model—another line of research focuses on designing inherently interpretable models.

In this work, we build on a matrix factorization model that is mathematically interpretable by design: user types are explicitly represented, and predicted scores are constrained to remain within the same range as observed ratings. This allows internal representations to be meaningfully interpreted in terms of user preferences. The challenge, however, lies in translating these internal representations into user-facing explanations.

Large language models (LLMs) offer a natural interface for generating such explanations. Given carefully designed prompts, they are capable of reasoning over structured information and expressing it in fluent natural language. This raises an important question: can LLMs successfully act as explanation generators for recommendation models whose internals are interpretable?

This paper investigates that question through three main objectives:

\begin{enumerate}
    \item \textbf{Evaluating an interpretable recommendation model.} We assess whether a mathematically interpretable model can satisfy users in terms of the recommendations it produces, even without additional explanation layers.
    
    \item \textbf{Generating explanations using LLMs.} We examine whether LLMs can leverage the model’s internal structure to generate effective and coherent explanations.
    
    \item \textbf{Comparing explanation strategies.} We contrast explanations grounded in model internals with alternatives based on external information (e.g., user history), to assess the trade-offs between transparency and other user-centered goals.
\end{enumerate}

These questions are addressed through a user study in which participants evaluate the recommendations and explanations they receive across multiple dimensions. Our findings highlight the potential of LLMs to act as a bridge between model interpretability and user-facing explanation, and offer guidance for designing explanation strategies that are both faithful and effective. All materials, including data preparation, prompts, and statistical analysis code, are available in a public repository to support transparency and reproducibility.\footnote{\url{https://github.com/MaximeUM/interpretable-mf-llm-explanations}}

\section{Related work}

Early research on explanations in recommender systems emphasized the importance of transparency and user-centered design. Herlocker et al.~\cite{herlocker2000explaining} conducted one of the first empirical studies on collaborative filtering (CF) explanation interfaces. They explored both white-box and black-box strategies and found that simple and transparent justifications—such as rating histograms and past accuracy—were more persuasive and trustworthy than abstract ones. Tintarev and Masthoff~\cite{tintarev2007survey} proposed a comprehensive framework for analyzing explanations, identifying seven core aims: transparency, scrutability, trust, effectiveness, persuasiveness, efficiency, and satisfaction. They highlighted the trade-offs between these objectives and the impact of explanation format and interface design.

Beyond system transparency, trust has emerged as a central but complex evaluation axis. Rong et al.~\cite{towards_explainable_ai} conducted a meta-analysis of 97 XAI user studies and showed that explanations generally improve subjective understanding and, to some extent, collaboration. However, their effects on trust and usability remain inconsistent. More targeted studies in recommender systems revealed similar complexity. For example, Ooge et al.~\cite{ooge2022trust} found that providing explanations—regardless of type—increased adolescents’ trust in educational recommendations.
Kunkel et al.~\cite{kunkel2019let} observed that personalized explanations boosted trust more effectively than impersonal ones. Millecamp et al.~\cite{millecamp2019music} further revealed that user traits (e.g., personality or domain expertise) moderated trust responses. Liao and Sundar~\cite{liao2021ai} highlighted framing effects, while Bucinca et al.~\cite{bucinca2021proxy} questioned the validity of proxy tasks for measuring trust, arguing for more holistic evaluations.

The emergence of large language models (LLMs) has significantly transformed explanation generation in recommender systems. Shi et al.~\cite{shi2024llm} proposed LLM-SRR, which enriches user reviews via LLMs and integrates them into knowledge graphs to identify explanatory paths. These are then translated into natural language, improving semantic faithfulness. Wang et al.~\cite{wang2025blessing} introduced LR-Recsys, a system using LLM-generated contrastive explanations embedded into DNN recommenders, which significantly improved performance. Gao et al.~\cite{gao2023chatrec} developed Chat-REC, a conversational recommender using structured prompts with LLMs to generate both recommendations and explanations.

Some works have directly evaluated user responses to LLM-generated explanations. Feng et al.~\cite{feng2025contextualizing}, through a user study, found that contextualized LLM explanations—those incorporating user history—improved users' intent to act and better met their cognitive needs compared to generic ones. 
Silva et al.~\cite{silva2024leveraging} compared personalized and generic explanations, showing that their effectiveness varies depending on the familiarity of the recommended item.

Several surveys now map the landscape of LLM-based explainable recommender systems. Zhao et al.~\cite{zhao2024recommender} categorized approaches by training paradigm (pre-training, fine-tuning, prompting) and noted the growing use of CoT prompting for nuanced reasoning. Lin et al.~\cite{lin2025how} proposed a 2D taxonomy (“where” and “how” to adapt LLMs), while Chen~\cite{chen2023survey} and Vats et al.~\cite{vats2024exploring} focused on explanation quality, personalization, and fairness. Across these surveys, a common theme is the lack of user-centered evaluation and the risk of hallucination in LLM explanations.

At the model level, several frameworks have pushed the boundaries of LLM alignment. Lei et al.~\cite{lei2024recexplainer} introduced RecExplainer, aligning LLMs with black-box recommenders through behavioral and intention-based strategies. Ma et al.~\cite{ma2024xrec} presented XRec, which uses collaborative embeddings from LightGCN and integrates them into LLMs via Mixture-of-Experts adapters. 
Bismay et al.~\cite{bismay2024reasoningrec} propose ReasoningRec, which uses LLM‑generated synthetic explanations to fine‑tune a smaller LLM, improving both recommendation accuracy and explanation quality.
Luo et al.~\cite{luo2024unlocking} developed LLMXRec, a two-stage approach using instruction-tuned LLMs for post-hoc explanation generation. Zhao et al.~\cite{zhao2024lane} introduced LANE, which uses Chain-of-Thought prompting and attention alignment to generate logical and transparent justifications.

Other notable contributions include consequence-based explanations~\cite{lubos2024improving}, which highlight the potential impact or outcomes of accepting a recommendation, and modular agent-based frameworks~\cite{peng2025survey}, which structure recommendation reasoning across profile, memory, planning, and action modules.

Despite these advances, most LLM-based methods focus on post-hoc explanations for black-box models, leaving a gap between formal interpretability and natural language justification. Our work addresses this gap by generating natural explanations from a matrix factorization model that is interpretable by design. To our knowledge, it is one of the first to bridge mathematical transparency and natural justification in a unified system.

\section{Methodology}

\subsection{Recommender system}

For the recommendation algorithm, we use BSSMF (Bounded simplex-structured matrix factorization) \cite{bssmf1,bssmf2}, which allows us to generate recommendations that are mathematically interpretable. This model decomposes the user-item interaction matrix \( X \in \mathbb{R}^{m \times n} \), where \( m \) is the number of items and \( n \) the number of users, into two smaller matrices, \( W \in \mathbb{R}^{m \times k} \) and \( H \in \mathbb{R}^{k \times n} \), such that \( X \approx WH \).

Each column of \( W \) represents a latent user type and contains the predicted ratings this latent type would assign to each item. Unlike traditional matrix factorization approaches where latent factors are unconstrained and often not directly interpretable, BSSMF contrains the values in \( W \) to lie within the same rating range as \( X \) (e.g., between 1 and 5). This makes it possible to interpret the values in \( W \) as actual scores, facilitating explanation.

Each column of \( H \) indicates how much a given user aligns with each latent type. BSSMF contrains the values in \( H \) to be non-negative and sum to one, meaning that each user is expressed as a convex combination of the latent user types--effectively forming a soft clustering over the user base.

Together, these constraints enable BSSMF to remain both expressive and inherently explainable: \( W \) reveals the preferences of interpretable user types, and \( H \) tells us how each user combines these types. This structure forms the basis for the explanations we generate throughout this study.
Moreover, BSSMF is much more robust to the choice of the fatorization rank, $r$, and to overfitting than standard unconstrained matrix factorization models~\cite{bssmf2}. 
The reason is that the additional bound constraints on $W$ and $H$ imply that the entries of $WH$ remain in the same range as the data.

\subsubsection{From ratings to recommendations}

To promote catalog diversity, we avoid recommending only the top-3 highest-scoring movies. Instead, we sample 3 recommendations from a pool of up to 20 candidate items with predicted scores $\geq 4$ (always including the actual top-3). The probability of selecting each movie is proportional to its predicted score. This approach balances relevance with diversity.

\begin{figure}[ht]
  \centering
  \includegraphics[width=\linewidth]{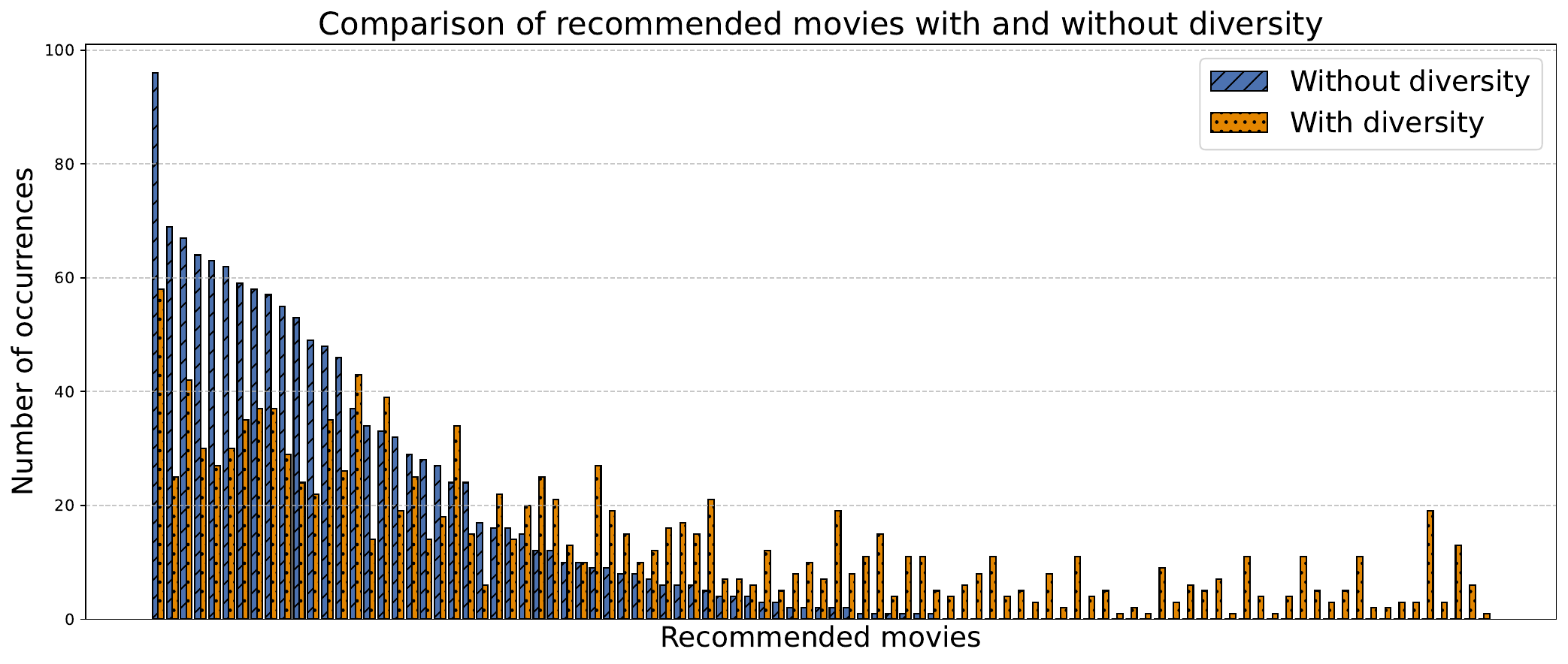}
  \caption{Comparison of the number of times each movie is recommended, with and without the sampling strategy.}
  \label{fig:recommended_diversity}
\end{figure}

As illustrated in Figure~\ref{fig:recommended_diversity}, this strategy significantly increases catalog coverage: the number of distinct recommended movies rises from 56 (with deterministic top-3) to 95 (with sampling), out of a set of 100 movies (see~\autoref{subsec:selec_movies} for details on the selection process).

\subsection{Generating explanations}
\label{methodology:generating_explanations}

As previously discussed, the BSSMF algorithm allows us to generate user types that can be interpreted. While this interpretation can be done manually for small-scale problems, it becomes infeasible as the data size increases—which is the case in most real-world recommender systems.

We therefore propose leveraging the capabilities of Large Language Models (LLMs) to:

\begin{enumerate}
    \item Interpret the user types. This serves two main purposes: (i)~providing interpretable insights into the latent user types, which are the building blocks used to reconstruct any individual user—valuable information for business teams, analysts, and system designers; and (ii)~reusing these interpretations to explain the recommendations made to users.
    \item Explain individual recommendations. By analyzing how a recommendation emerges from the weighted combination of user types, we can generate a natural language explanation tailored to each user and item.
\end{enumerate}

\subsubsection{Generating explanations in practice}

To generate meaningful explanations using a Large Language Model (LLM), two elements are critical: selecting a model with sufficient reasoning capabilities, and designing a prompt that aligns with the structure of the data. We evaluated several instruction-tuned models, including Llama 3 3B Instruct, Llama 3 8B Instruct \cite{llama3}, Mixtral 8x7B Instruct, Mixtral 8x22B Instruct \cite{mixtral}, Mistral Small 24B Instruct, Llama 3 70B Instruct \cite{llama3}, and DeepSeek R1 Distill Llama 70B \cite{deepseekai2025}.

For each model, we assessed two aspects: the quality of the descriptions generated for user types (e.g., are they well-defined, coherent, and distinctive?) and the quality of the explanations provided for individual users. Among the tested models, DeepSeek consistently delivered the best results on both fronts. Its reasoning is more structured, and it returns its step-by-step thought process before delivering the final answer. This feature is particularly valuable for us, as it allows us to better understand how the model arrives at a given explanation—facilitating post hoc analysis and qualitative validation of its outputs.

Although we do not detail this model comparison process here---since comparing LLMs is not the main goal of this paper---it required considerable effort and tuning. We emphasize that the objective is not to benchmark models, but to verify that a sufficiently capable LLM can generate high-quality, interpretable explanations within our setup.

The selected model, DeepSeek-R1-Distill-Llama-70B, is not the largest in terms of parameter count, but it is nonetheless substantial. Running it in inference mode with float16 precision requires a setup with eight A40 GPUs (40GB VRAM each).

\paragraph{Explaining the user types}
\label{explain_user_types}

The first step in our pipeline is to interpret the latent user types obtained from our matrix factorization model. We prompt the LLM with the following instruction:

\vspace{0.5em}
\begin{framed}
\normalsize\ttfamily
\noindent
We have a recommendation system based on matrix factorization, capable of generating recommendations. We aim to interpret the user types. 
The matrix X (containing ratings between 1 and 5) has dimensions (movies x users). We decompose it as follows: X = W * H, where W has dimensions (movies x latent factors) and H has dimensions (latent factors x users). 
A rating above 4 is considered very good. A rating of 3 is acceptable. A rating below 3 indicates less interesting movies for the latent user. 
The values in W fall within the same range as those in X. Each column of W represents a user type. Each column of H sums up to 1.  
Your role is to interpret the user types. 
For each user type, provide a description in a maximum of 100 words. Base your explanation on all the movies associated with the user type. 
It is very important that the descriptions of the user types are explicit and different. 
Each user type's description should neither be too obvious nor too generalistic, in order to obtain distinctive and characteristic user types. 
We provide you with a column of W. Do not list too many liked movies to describe the user type, but rather focus on the characteristics of these movies... You can mention 2–3 movies if necessary, but do not make a long list of liked movies. 
Be careful not to say that a user type likes a movie if that movie has a rating below 4. 
If you realize that your descriptions for two (or more) user types are too similar, it means you are not distinguishing them enough. Focus on what really differentiates them. 
Please reason step by step, and put your final answer within \textbackslash boxed\{\}.
\end{framed}
\vspace{0.5em}

This prompt was carefully designed to guide the LLM toward producing compact, distinct, and insightful descriptions of each latent user type, while enforcing consistency with the underlying data.
\\\\
To build the user prompt, we extract each column of the matrix \( W \), representing the preferences of a user type across all movies. Since a user type is mathematically defined by its scores over the entire item space, we include all movies, ranked by predicted rating. For each, we provide the French title, predicted score, and translated genres—enabling the LLM to interpret the user type meaningfully, as item IDs alone would not be understandable. This results in a purely mathematical interpretation made accessible through natural language.

\paragraph{Explaining the recommendations}

To explain the recommendations, we consider several approaches.

The first, already introduced, is to leverage the mathematical structure of the model to generate natural language explanations. In this approach, we use the latent user types and the model internal weighting to justify recommendations in a transparent and faithful way.

\newpage
\vspace{0.5em}
\noindent\textbf{Model-based explanation (user types + latent weights):}
\begin{framed}
\normalsize\ttfamily
\noindent
We use a recommendation system based on matrix factorization to suggest relevant movies based on user preferences.
\\
        How it works:\\
      \tabindent  - The matrix X (movies x users) contains ratings from 1  to 5.
      
        - We decompose it as X = W * H, where:
        
        \tabindent - W (movies x latent factors) represents ratings from user types.
        
        \tabindent - H (latent factors x users) weights the influence of each user type for \tabindent \tabindent a given user (each colum of H sum up to 1).
        
        - \textbf{A movie's final score is a weighted average of evaluations from \tabindent multiple user types.}
\\
        Your task:
        
        - Explain why this movie might appeal to the user \textbf{without mentioning \tabindent matrix factorization or user type weightings.}
        
        - Highlight broad trends rather than linking the recommendation to a single \tabindent user type.
\\
        Guidelines:
        
        - Justify the recommendation in \textbf{a maximum of two sentences}.
        
        - Emphasize \textbf{thematic, tonal, stylistic, or emotional similarities} \tabindent rather than just genre overlap.
        
        - Adopt a \textbf{natural, professional, and engaging tone}.
        
        - Frame the explanation as \textbf{insightful advice, not a film summary}.
        
        - Avoid robotic or generic phrasing.
        
        - \textbf{Express measured enthusiasm} to spark interest without exaggeration.
        
        - Please reason step by step, and put your final answer within \textbackslash boxed\{\}.
\\
        Final Goal:
        
        - The recommendation should feel meaningful, not generic.
        
        - The explanation should spark curiosity and interest.
        
        - \makebox[0pt][l]{\textbf{Reflect a mix of preferences, not a single user type’s}}

        \makebox[0pt][l]{\textbf{perspective.}}
\\
        Finally, translate everything into French and use the informal "tu" form. At the end, we only want the French final answer in the box. Do not add information, just the final answer.
\end{framed}

\vspace{0.5em}

We also explore an alternative strategy that relies on simplified information: instead of referring to the model’s internal computations, the explanation is based on the user’s previously liked movies. While this may seem intuitive, it is less transparent, as it does not reflect how the model actually works. Our recommender system is based on BSSMF which embeds users and items in a shared latent space and makes recommendations via dot products between these embeddings. As a result, the model does not directly rely on previously liked items when recommending new ones. Explanations based on viewing history, although easy to understand, do not faithfully represent the reasoning behind the recommendations.

\vspace{0.5em}
\noindent\textbf{History-based explanation (based on liked movies):}
\begin{framed}
\normalsize\ttfamily
\noindent We generate a personalized explanation to help the user understand why a recommended movie might be a good match.
\\
        How it works:
        
        - The explanation is based on:
        
          \tabindent - The \textbf{title} of the recommended movie.
          
          \tabindent - Its \textbf{genres}.
          
          \tabindent - \makebox[0pt][l]{The \textbf{titles and genres of movies the user has previously}} 
          
         \tabindent \textbf{watched and rated highly (at least 4 stars).}

        - The goal is to \textbf{highlight meaningful connections} between past \tabindent preferences and this recommendation, without mentioning an algorithm.
\\
        Your task:
        
        - Justify the recommendation in \textbf{a maximum of two sentences}.
        
        - Emphasize \textbf{thematic, tonal, stylistic, or emotional similarities} \tabindent rather than just genre overlap.
        
        - Adopt a \textbf{natural, professional, and engaging tone}.
        
        - Frame the explanation as \textbf{insightful advice, not a film summary}.
        
        - Avoid robotic or generic phrasing.
        
        - \textbf{Express measured enthusiasm} to spark interest without exaggeration.
        
        - Please reason step by step, and put your final answer within \textbackslash boxed\{\}.
\\
        Final goal:
        
        - The explanation should feel \textbf{relevant and meaningful} to the user.
        
        - It should \textbf{spark curiosity and encourage them to watch the movie}.
\\
        Finally, translate everything into French and use the informal "tu" form. At the end, we only want the French final answer in the box. Do not add information, just the final answer.
\end{framed}

\vspace{0.5em}

Finally, we include a third type of explanation that combines both previous approaches: it references the user types and the model’s logic, while also connecting the recommended item to the user’s past preferences. This hybrid strategy aims to benefit from both transparency and familiarity.

\vspace{0.5em}
\noindent\textbf{Combined explanation (model reasoning + history):}
\begin{framed}
\normalsize\ttfamily
\noindent We use a recommendation system based on matrix factorization to suggest relevant movies based on user preferences.
\\
        How it works:
        
        - The matrix X (movies x users) contains ratings from 1 to 5.
        
        - We decompose it as X = W * H, where:
        
         \tabindent - W (movies x latent factors) represents ratings from user types.
          
         \tabindent - H (latent factors x users) weights the influence of each user type for \tabindent \tabindent a given user (each column of H sums up to 1).
         
        - \textbf{A movie's final score is a weighted average of evaluations from \tabindent multiple user types.}
        
        - In addition to this approach, we consider the user's \textbf{history of highly \tabindent rated movies (score $\geq$ 4)} to refine the explanation.
        
        - This means the reasoning should not only reflect general trends from \tabindent user types but also highlight \textbf{connections with movies the user has \tabindent previously enjoyed}.
\\
        Your task:
        
        - Explain why this movie might appeal to the user \textbf{without mentioning \tabindent matrix factorization or user type weightings}.
        
        - Justify the recommendation in \textbf{a maximum of two sentences}.
        
        - Emphasize \textbf{thematic, tonal, stylistic, or emotional similarities} \tabindent rather than just genre overlap.
        
        - When possible, draw meaningful \textbf{parallels with movies the user has \tabindent highly rated} to reinforce the recommendation.
        
        - Adopt a \textbf{natural, professional, and engaging tone}.
        
        - Frame the explanation as \textbf{insightful advice, not a film summary}.
        
        - Avoid robotic or generic phrasing.
        
        - \textbf{Express measured enthusiasm} to spark interest without exaggeration.
        
        - Please reason step by step, and put your final answer within \textbackslash boxed\{\}.
\\
        Final goal:
        
        - The explanation should feel \textbf{relevant and meaningful} to the user.
        
        - It should \textbf{spark curiosity and encourage them to watch the movie}.
        
        - \makebox[0pt][l]{\textbf{Reflect a mix of preferences, not a single user type’s}} 
        
        \tabindent \textbf{perspective} while incorporating insights from the user's past interests.
\\
        Finally, translate everything into French and use the informal "tu" form. At the end, we only want the French final answer in the box. Do not add information, just the final answer.
\end{framed}

\section{Study Design}

\subsection{Selecting a representative and usable movie subset}
\label{subsec:selec_movies}

Choosing an appropriate dataset for a user study in recommender systems is not trivial. While most benchmarks rely on well-known datasets such as MovieLens 100K or 1M \cite{harper2015movielens}, these collections contain relatively old movies, which may not reflect the type of content typically consumed by modern users. To address this limitation, we opted to extract a subset of movies from the larger MovieLens 32M dataset\footnote{\url{https://grouplens.org/datasets/movielens/32m/}}.

Our decision to work with a reduced subset was not due to computational constraints—BSSMF scales to larger datasets—but rather because the number of user interactions we expect to collect is limited. A smaller, well-curated matrix allows us to preserve the structure of a realistic recommendation scenario while keeping the study manageable.

To build a representative and diverse pool of 100 movies for the user study, we selected items based on popularity while penalizing older movies to avoid temporal bias. French titles, synopses, and poster images were retrieved using the TMDB API\footnote{\url{https://developer.themoviedb.org/}} to ensure language consistency. We also ensured diversity by limiting the selection to one movie per saga. The full data preparation pipeline is available in our GitHub repository.

\subsection{Selecting a consistent user base}

To collect user data, we relied on accessible participants to whom we could easily distribute the questionnaire. Specifically, we collected responses from students and staff members at our university. Since the university is French-speaking, the entire study was conducted in French to match participants’ native language. While this introduces a potential language-related bias, we consider it minimal—such a bias would also exist if the study were conducted in English.

We are aware of the sampling bias that results from recruiting participants from an academic environment, as it restricts our sample to a relatively educated segment of the population. However, we argue that since movies are a widely consumed cultural product, this bias is mitigated in practice and does not prevent us from collecting coherent and meaningful data for our purposes.

In total, we collected data from 326 users (see \autoref{subsec:collecting_data} for details).

\subsection{Collecting data}
\label{subsec:collecting_data}

As we aim to conduct a user-centered study, it is essential to define a rigorous data collection protocol. To guide our methodology, we follow best practices outlined in \cite{towards_explainable_ai}, which provides a comprehensive overview of experimental designs for explainability-focused user studies.

\subsubsection{Collecting ratings to train the recommendation algorithm}

To train the recommendation algorithm, we collected 10 ratings from each of 440 participants, using the curated pool of 100 movies. Each participant rated a random subset of 10 movies, promoting matrix diversity. A synopsis (retrieved via the TMDB API\footnote{\url{https://developer.themoviedb.org/}}) was provided when needed. Users rated each movie using a 5-point Likert scale: \textit{“I really like it”, “I like it”, “It's okay”, “I don't like it much”, “I really don't like it”}. This resulted in 4400 user-item interactions. \label{likert_films}

\subsubsection{Evaluating recommendations and explanations}
\label{evaluating_recommendations_and_explanations}

To evaluate the usefulness and impact of explanations, we employ a between-subjects design, dividing users into four distinct groups, each exposed to a different type of explanation (or none). This decision is supported by findings in \cite{towards_explainable_ai}, which reports that 55\% of reviewed studies adopt a between-subjects design. While within-subjects designs allow each participant to compare different conditions—e.g., by ranking explanations—they introduce significant biases. In particular, asking participants to compare explanations implicitly suggests that some explanations must be better than others, thereby confounding the evaluation of whether explanations actually enhance understanding or trust. In contrast, the between-subjects design allows us to evaluate the intrinsic value of each explanation type without introducing comparative framing effects.

We split participants into four groups depending on the explanation strategy: one group receives only the recommendation and synopsis (no explanation), while the other three receive model-based, history-based, or combined explanations, as detailed in \autoref{methodology:generating_explanations}.

The 440 participants from the initial rating phase were evenly assigned to one of the experimental groups for the recommendation evaluation. In total, 326 participants completed this second phase, each evaluating three personalized recommendations, for a total of 978 evaluations.

For each recommended movie, participants rated the recommendation (U2) and had the option to leave a free-text comment.

Participants in Groups~1, 2, and 3 (who received explanations) also evaluated the explanation across several key dimensions using a 5-point Likert scale: 
\textit{“Strongly agree”, “Somewhat agree”, “Neither agree nor disagree”, “Somewhat disagree”, “Strongly disagree”}.

To avoid bias, the statements were presented in random order. The evaluated dimensions and their associated statements are summarized in \autoref{dimensions_and_questions}.

\begin{table}[ht]
\caption{Evaluation questions and their corresponding target dimensions. U2 is for all groups; the others apply to Groups 1, 2, and 3.}
\centering
\begin{tabular}{p{0.26\linewidth}@{\hskip 6pt}p{0.68\linewidth}}
\toprule
\textbf{Dimension} & \textbf{Associated statement(s)} \\
\midrule
\textit{Recommendation evaluation} &
\textbf{U2}: What is your opinion of this movie? (Rated on the 5-point Likert scale from \autoref{likert_films}) \\
\midrule
Transparency &
\textbf{T1}: The explanation helps me understand why this movie was recommended to me. \newline
\textbf{T2}: The explanation allows me to understand, in broad terms, how the recommendation system works. \\
Effectiveness &
\textbf{E1}: Thanks to this explanation, I feel I can make an informed decision about whether I will like this movie or not. \newline
\textbf{E2}: The explanation helps me determine whether I would like the recommended movie. \\
Persuasion &
\textbf{P1}: The explanation makes me more likely to follow the recommendation and watch the movie. \\
Trust &
\textbf{TR1}: Thanks to the explanation, I trust the system more to recommend movies that match my tastes. \\
Satisfaction &
\textbf{S1}: Overall, I am satisfied with the provided explanation. \\
\bottomrule
\end{tabular}

\label{dimensions_and_questions}
\end{table}

We included two items for both \textit{transparency} and \textit{effectiveness}, as these dimensions are conceptually broader. \textbf{T1} and \textbf{T2} capture local (why this item) and global (how the system works) understanding, respectively. \textbf{E1} assesses whether the explanation supports informed decision-making, while \textbf{E2} targets intuitive alignment with user preferences. The other dimensions—\textit{persuasion}, \textit{trust}, and \textit{satisfaction}—are each measured by a single, focused item.

\paragraph{Impact of explanations on recommendation evaluation} A key question in explainable recommendation is whether explanations can enhance users' appreciation of the recommendations themselves. This goes beyond transparency or trust to ask whether users actually like the recommendations more when they are explained. 

This hypothesis is rooted in classical psychology: the \textit{because effect}~\cite{langer1978mindlessness} shows that people are more likely to comply with a request when given a reason—even if it is trivial. However, the notion of ``acceptance'' can be ambiguous: does an explanation make users more likely to engage with the content, or does it bias their evaluation of the item?

Our design allows us to investigate this question via question \textbf{U2}, which is asked in all groups, including the one with no explanation. Comparing the U2 scores between Group~0 (no explanation) and Groups~1–3 (with explanation) provides an estimate of whether receiving an explanation affects users’ appreciation of the recommendation. 

\subsection{Experimental setup}

We base our study on a BSSMF model trained on a subset of the MovieLens 32M dataset. The interaction matrix includes $n = 135{,}616$ users, which comprises both MovieLens users and the $440$ real users who participated in our evaluation. A total of $100$ items and $3{,}898{,}839$ user-item interactions are retained after filtering.

To evaluate generalization and guide the choice of latent dimensionality $r$, we split the dataset into a training and a test set. We randomly select $5$ interactions per user for the test set, restricted to users with more than $10$ ratings. The remaining interactions are used for training. 

To guide the choice of latent dimensionality \(r\), we trained models with \(r = 3\), \(5\), and \(10\). As expected, increasing \(r\) improved training accuracy (RMSE = 0.72, 0.68, and 0.61 respectively), but test RMSE did not follow the same trend: 0.83 for \(r=3\), 0.85 for \(r=5\), and 0.89 for \(r=10\). We selected \(r=5\) as a compromise, balancing generalization and interpretability, with a manageable number of distinct and meaningful user types.

\section{Results}

\subsection{Discovered user types}

The method described in \autoref{explain_user_types} allows us to uncover the user types that characterize our system. In our setup, we identify the following user types:

\let\FrameCommand\OldFrameCommand 
\begin{framed}
\normalsize\ttfamily
\noindent
\textbf{User Type 1: The Epic Enthusiast}  

\noindent This user craves high-octane, emotionally charged experiences. They favor intense, dramatic films like "Saving Private Ryan" and "Braveheart," which offer epic storytelling and serious themes. While they enjoy a mix of genres, including lighter fare like "Zootopia", their true passion lies in dramatic, action-packed narratives that leave a lasting impact.
\\\\
\textbf{User Type 2: The Thought Provoker}
\\
With a penchant for dark, psychological themes, this user seeks films that challenge their perspective. Movies like "Parasite" and "Fight Club" reveal a love for complex, thought-provoking narratives that delve into the human condition, offering both intellectual stimulation and emotional depth.
\\\\
\textbf{User Type 3: The Nostalgic Visionary}  
\\
This user appreciates a blend of nostalgia and visual artistry. Films like "Aladdin" and "Twelve Monkeys" showcase their enjoyment of both timeless stories and visually stunning cinema. They value strong narratives and memorable visuals, savoring a diverse range of genres but always seeking that special cinematic magic.
\\\\
\textbf{User Type 4: The Thrill Seeker}  
\\
Encompassing a broad spectrum of genres, this user thrives on high-energy experiences. From action-packed blockbusters like "Terminator" to suspenseful dramas like "Gone Girl", they love stories with complex characters and thrilling plots, always chasing the next adrenaline rush in their cinematic journey.
\\\\
\textbf{User Type 5: The Dark Explorer}  
\\
This user is drawn to intense, unconventional narratives. Films like "Joker" and "Pulp Fiction" highlight their taste for dark, boundary-pushing themes. They enjoy both action and intellectual engagement, often seeking stories that explore the deeper, darker aspects of human nature and society.
\end{framed}

\subsection{Examples of explanations}

To illustrate the nature of the generated explanations, \autoref{tab:example_explanations} presents three representative examples per group, each reflecting the typical style and content characteristic of its explanation category.

\begin{table}[ht]
\caption{Examples of generated explanations per group.}
\label{tab:example_explanations}
\centering
\begin{tabular}{@{}p{0.15\linewidth}p{0.82\linewidth}@{}}
\toprule
\textbf{Group} & \textbf{Explanation} \\
\midrule

Group 1 & 
\begin{minipage}[t]{\linewidth}
\textit{Forrest Gump might interest you as it blends an epic and emotional story, touching on key historical moments, with deep characters and surprising twists.}

\vspace{0.4em}\hrule\vspace{0.4em}

\textit{You'll love ``Jurassic Park'' for its mix of epic action and deeper reflections on human ambition, all brought to life with stunning visuals.}

\vspace{0.4em}\hrule\vspace{0.4em}

\textit{``Iron Man'' combines high-octane action with thoughtful storytelling, making it an exciting choice for those who enjoy both thrilling adventures and stories that explore deep themes such as identity and responsibility.}
\end{minipage}
\\
\midrule

Group 2 & 
\begin{minipage}[t]{\linewidth}
\textit{You’re going to love ``The Martian'' because, like ``Guardians of the Galaxy'' or ``Interstellar'', it perfectly blends space adventure with a moving, hope-filled story. The film captures that balance between suspense and emotion that resonated with fans of ``Good Will Hunting'' and intense dramas.}

\vspace{0.4em}\hrule\vspace{0.4em}

\textit{You might enjoy ``Iron Man'' because it blends action and intelligence, with a complex character and deeper themes, much like ``Gone Girl'' or ``Batman''. It’s also a visually stunning adventure — perfect for those who liked ``Thor'' or ``Jurassic Park''.}

\vspace{0.4em}\hrule\vspace{0.4em}

\textit{``Gone Girl'' might appeal to you because, like ``Ocean's Eleven'' and ``Ex Machina'', it pulls you into a twisted and psychological story with gripping twists. Its mix of suspense and reflection on human relationships might spark your curiosity, just like the intense films you've enjoyed.}
\end{minipage}
\\
\midrule

Group 3 & 
\begin{minipage}[t]{\linewidth}
\textit{You might enjoy this film as, like ``Se7en'' or ``Memento'', it cleverly blends suspense, a complex plot, and striking revelations with strong psychological tension.} 

\vspace{0.4em}\hrule\vspace{0.4em}

\textit{This movie might interest you because it combines a gripping story with intellectual stakes, much like ``The Bourne Identity''. The emotion and depth of the characters, similar to those in ``Forrest Gump'', will draw you into a historical and suspenseful narrative.}

\vspace{0.4em}\hrule\vspace{0.4em}

\textit{You'll love ``Django Unchained'' for its mix of intense action and deep dramatic storytelling, much like in ``Saving Private Ryan''. This film, with its complex characters and reflection on dark themes, will also remind you of ``The Silence of the Lambs'' and ``American History X'', with a unique style that will captivate you.}
\end{minipage}
\\
\bottomrule
\end{tabular}
\end{table}

\subsection{How are the explanations and recommendations perceived overall?}

We begin by analyzing how users globally perceive the recommendations and explanations they received, regardless of the specific evaluation dimensions defined earlier. The goal here is to obtain an overall picture of user responses across the different experimental conditions.

To this end, we compute and compare the mean scores for each evaluation question, across all four groups defined in our study. This allows us to assess how each type of explanation influences the way users experience the system.

\begin{figure}[ht]
  \centering
  \includegraphics[width=\linewidth]{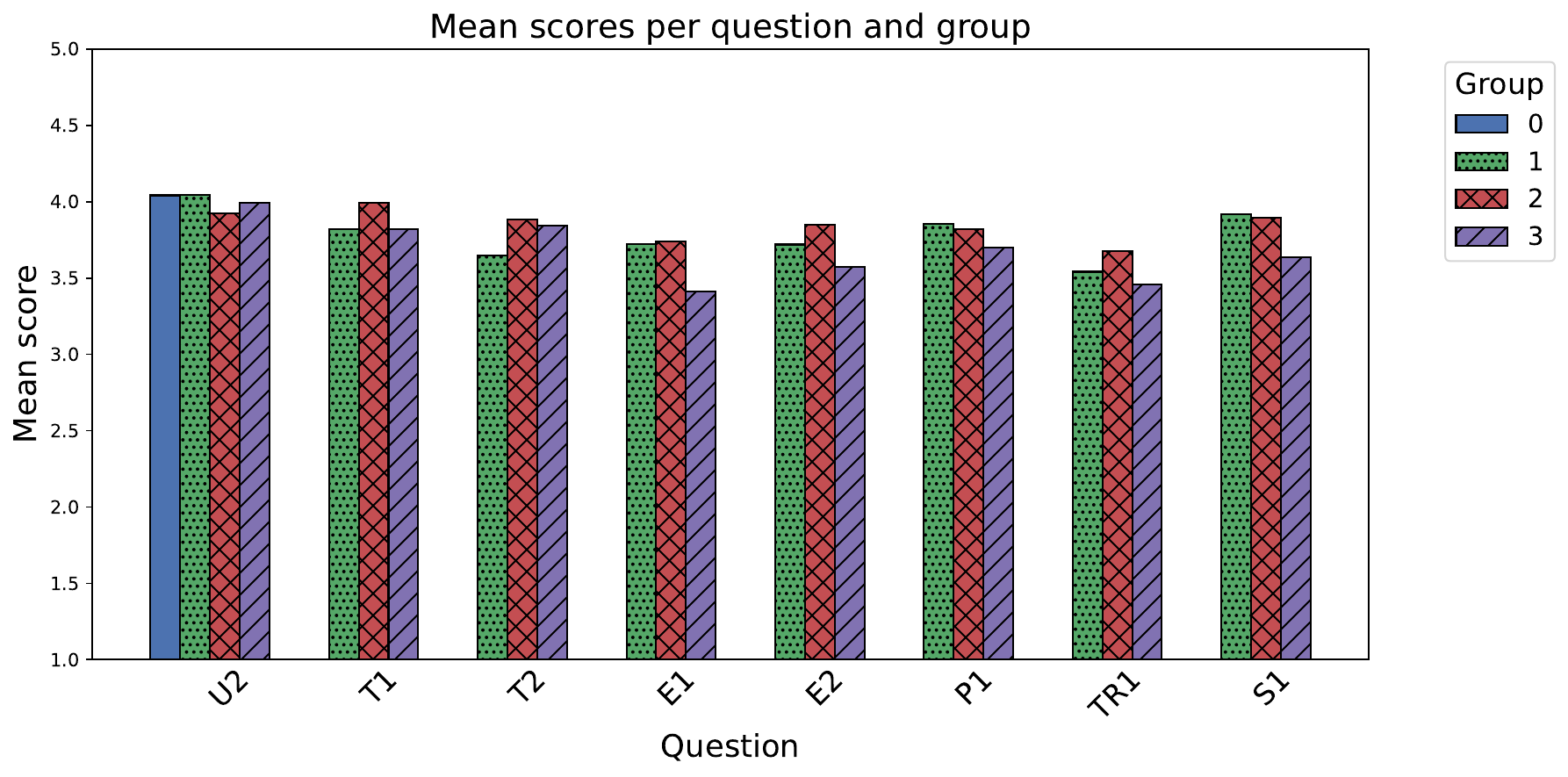}
  \caption{Mean questions scores across the four experimental conditions.}
  \label{fig:mean_scores}
\end{figure}

\autoref{fig:mean_scores} shows the mean scores across all evaluation criteria, indicating consistently high satisfaction levels in all groups. Since the median score is 4 for every group and criterion, we do not report medians separately.
In addition to mean scores, we also analyzed alternative indicators such as the proportion of low ratings ($\leq$ 2), percentile distributions, and raincloud plots. As these analyses yielded results consistent with the comparison of means, we opted to present this more compact and synthetic figure. All complementary analyses and visualizations are available in our GitHub repository.

\subsection{Statistical analysis}

A common approach for comparing multiple groups is the one-way ANOVA test~\cite{towards_explainable_ai}, which assumes normality, homoscedasticity, and independent observations. While independence is guaranteed by our experimental design (random group assignment), the other assumptions must be tested. We use the Shapiro–Wilk test \cite{shaphiro1965analysis} for normality and Levene’s test \cite{levene1960robust} for homoscedasticity.

As expected given the limited range of Likert responses, normality is clearly violated ($p = 0.000$ in all cases). Homoscedasticity is only partially satisfied, with Levene’s $p$-values ranging from $0.001$ to $0.286$. Although ANOVA is sometimes used despite such violations, we adopt a more robust approach and rely on the Kruskal–Wallis test \cite{kruskal1952use, mckight2010kruskal}—its non-parametric equivalent—which does not assume normality nor homoscedasticity.
Results are reported in \autoref{kruskal-wallis}, and full details of the tests are provided in our GitHub repository.

\begin{table}[ht]
\caption{Kruskal–Wallis test results across user groups}
\label{kruskal-wallis}
\begin{tabular}{lrrl}
\toprule
 Question Abbr.   &   H statistic &   p-value & Interpretation   \\
\midrule
 U2               &       1.67071 &     0.643 & Not significant  \\
 T1               &       6.38245 &     0.041 & \textbf{Significant}      \\
 T2               &       7.20286 &     0.027 & \textbf{Significant}      \\
 E1               &      13.1742  &     0.001 & \textbf{Significant}      \\
 E2               &       9.53791 &     0.008 & \textbf{Significant}      \\
 P1               &       3.13838 &     0.208 & Not significant  \\
 TR1              &       5.10687 &     0.078 & Not significant  \\
 S1               &      14.3696  &     0.001 & \textbf{Significant}      \\
\bottomrule
\end{tabular}

\end{table}

\subsubsection{Impact of the explanation on recommendation appreciation}

As shown in \autoref{kruskal-wallis}, we observe no statistically significant differences between groups for U2. This suggests that explanations do not bias users’ stated preferences toward the recommended items, consistent with findings from Lu et al.~\cite{Lu2023}, who found only marginal changes in preference ratings before and after exposure to explanations—except when explanations were provided by peers.

However, as shown in \autoref{fig:mean_scores}, among the participants who received an explanation (Groups~1--3), responses to \textbf{P1} (persuasion) indicate that users are generally more inclined to follow the recommendation. This suggests that while explanations do not significantly affect the stated preference for the movie (U2), they may increase users’ willingness to engage with the recommended content.

\subsubsection{Perception of the explanations across experimental groups}

\label{subsec:perception_explanations}

To identify which specific explanation strategies differ, we perform post hoc pairwise comparisons using Dunn’s test~\cite{dunn1964multiple}, a non-parametric method suited for rank-based data. The test is applied only to questions where a significant group effect was detected. Results are summarized in \autoref{dunn}.

\begin{table}[ht]
\caption{Post hoc pairwise comparisons between groups using Dunn’s test}
\label{dunn}
\begin{tabular}{lrrrl}
\toprule
 Question   &   Group A &   Group B &   p-value & Interpretation   \\
\midrule
 T1         &         1 &         2 &    0.0369 & \textbf{Significant}      \\
            &         1 &         3 &    1      & Not significant  \\
            &         2 &         3 &    0.2756 & Not significant  \\
 T2         &         1 &         2 &    0.0441 & \textbf{Significant}      \\
            &         1 &         3 &    0.1036 & Not significant  \\
            &         2 &         3 &    1      & Not significant  \\
 E1         &         1 &         2 &    1      & Not significant  \\
            &         1 &         3 &    0.0083 & \textbf{Significant}      \\
            &         2 &         3 &    0.0037 & \textbf{Significant}      \\
 E2         &         1 &         2 &    0.442  & Not significant  \\
            &         1 &         3 &    0.2369 & Not significant  \\
            &         2 &         3 &    0.0064 & \textbf{Significant}      \\
 S1         &         1 &         2 &    1      & Not significant  \\
            &         1 &         3 &    0.0009 & \textbf{Significant}      \\
            &         2 &         3 &    0.0176 & \textbf{Significant}      \\
\bottomrule
\end{tabular}

\end{table}

\begin{figure}[ht]
  \centering
  \includegraphics[width=\linewidth]{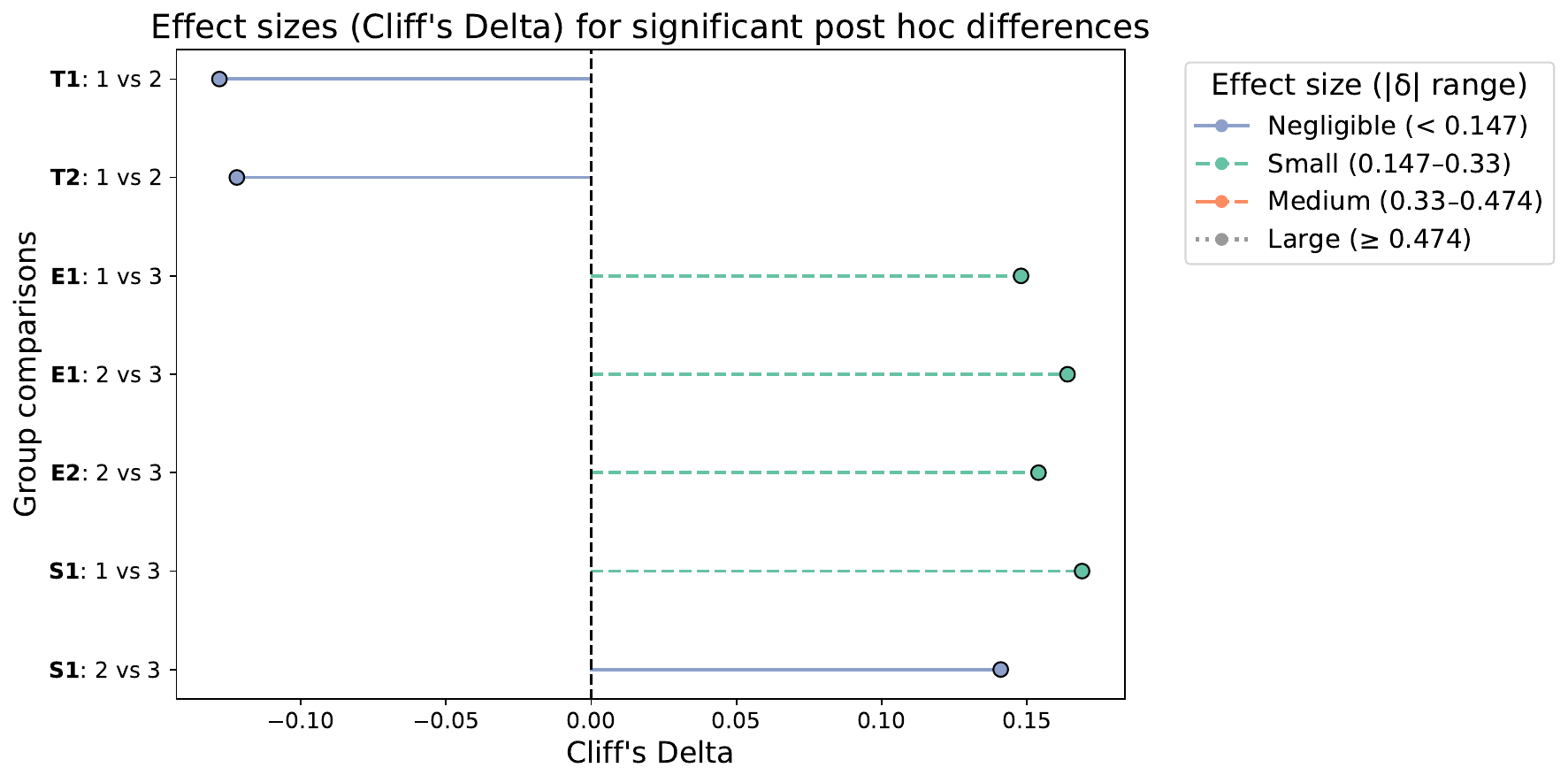}
  \caption{Cliff’s delta for all significantly different pairs identified by Dunn’s test. Values indicate both the direction and magnitude of the effect between groups.}
  \label{fig:cliff}
\end{figure}

While Dunn’s test identifies statistically significant differences, it does not indicate the direction or magnitude of these differences. We therefore compute Cliff’s delta~\cite{meissel2024using,grissom2012effect} for each significant pairwise comparison. This non-parametric effect size measures how frequently values from one group exceed those from another, ranging from \(-1\) (complete dominance of Group~B) to \(+1\) (complete dominance of Group~A), with 0 indicating no difference. The results are visualized in \autoref{fig:cliff}.

\paragraph{Interpretation of results.}

Effect sizes computed via Cliff’s delta (\autoref{fig:cliff}) show that all significant differences between groups fall within the negligible or small range. This suggests that while certain explanation strategies are rated differently, these differences remain modest in practice—reinforcing that all approaches are generally well received.

For the transparency dimension (T1 and T2), Group~2 is perceived as more transparent than Group~1. This is somewhat counterintuitive, as Group~1 explanations are grounded in the model’s internal logic, whereas Group~2 explanations rely solely on the user’s viewing history. However, qualitative feedback suggests that Group~2 explanations may feel more relatable, thanks to their explicit references to familiar movies—even if they are not faithful to the actual reasoning of the model.

Regarding effectiveness, Group~3 is rated less favorably than Groups~1 and~2 for E1 (informativeness), and less favorably than Group~2 for E2 (ability to judge whether one would like the movie). Although Group~3 explanations combine internal reasoning with historical context, this hybrid strategy may introduce verbosity or less coherent narratives. Some users expressed (see \autoref{qualitative_feedback}) confusion over the links drawn between movies and a desire for clearer, more focused justifications.

For satisfaction (S1), Group~3 also receives slightly lower ratings compared to Groups~1 and~2. This suggests that richer explanations do not always translate to greater satisfaction—possibly due to information overload or a lack of narrative clarity.

Overall, these findings highlight a trade-off: while combining different information sources can enrich explanations, simpler and more targeted strategies may prove more effective if they better match user expectations and maintain clarity.

\subsection{Qualitative feedback} 
\label{qualitative_feedback}
We analyzed the free-text comments provided by users in each condition to better understand how explanations were perceived. As expected, participants in Group~0, who received no explanation, mostly left short confirmations that the recommendations were appealing, but their comments lacked further elaboration. In Group~1 (explanations based on model internals), several users appreciated the presence of an explanation and even stated that it made them more interested in the movie than the synopsis itself. However, many found the wording overly generic, emotional, or vague, with comments suggesting that the explanations could apply to any movie or any user. Some participants noted a lack of concrete elements or specificity, and a few were bothered by stylistic choices (e.g., being addressed informally). Group~2 (history-based explanations) generated more polarized feedback: participants liked the references to previously seen movies, but often criticized the lack of depth in the justification, the repetitive nature of some phrases, or the weak relevance of the mentioned titles. The frequent reuse of the same reference movies—caused in our setup by the limited number of previously rated movies (10 per user)—was noted as a limitation. This reinforces the idea that history-based explanations may be less suitable in cold-start scenarios, where user interaction data is sparse and diversity in justifications is harder to achieve. In Group~3 (hybrid explanations), several users struggled to understand the connections made between recommended movies and their viewing history, sometimes pointing out that the movies seemed unrelated or that the links felt unjustified. Overall, while each explanation strategy had its strengths, user comments highlighted recurrent concerns about vagueness, genericity, and the need for more specific and grounded justifications.

\section{Conclusion and Future Work}

This work investigates how interpretability and natural language explanations can be jointly integrated into recommender systems in a way that is both faithful to the underlying model and understandable to users. We propose a two-stage framework that starts from a constrained matrix factorization algorithm—BSSMF—designed to yield human-interpretable latent factors, and then translate these signals into textual justifications using a large language model. Unlike many existing approaches that generate explanations post hoc for black-box models, our method grounds the generation process in a model whose behavior is fully transparent and mathematically interpretable.

Through a controlled user study, we compared four explanation strategies—including no explanation, LLM-based explanations faithful to the model’s structure, explanations based on user history, and a combination of both. Results show that even without explanations, users consistently rated the recommendations highly, suggesting that interpretable models such as BSSMF are able to provide strong relevance signals by design. This supports the idea that interpretability and performance can coexist and that models crafted with transparency in mind can serve as solid foundations for explainable AI.

However, our study also reveals important nuances. Although strategy~1 (based on the model’s internal logic) is the most faithful to how recommendations are actually computed, it was not always rated higher than strategy~2 (based on user history), which some users found more familiar or personally meaningful. This points to a fundamental challenge in explainable recommendation: the explanations that are most truthful are not always the most effective in terms of user perception. The design of future explanations should take into account transparency, effectiveness, persuasion, trust, and overall user satisfaction, along with other factors that may influence the user experience.

Furthermore, the movie domain naturally provides a wealth of contextual signals—e.g., visual elements, familiar titles, and synopses—that may reduce users' reliance on explanations. The same system deployed in domains such as job recommendation, online education, or medical decision support could yield very different outcomes. Explanations in those contexts are likely to be more critical for building user trust, ensuring fairness, and supporting informed decision-making. Generalizing our approach to such domains will be an important direction for future research.

Another insight from our qualitative analysis is that users often perceive the explanations as repetitive. Despite the LLM’s fluency and coherence, repeated exposure to similar explanations reduces their perceived usefulness. This highlights the importance of introducing diversity into explanation content, to sustain user engagement over time.
In parallel, we observe that users differ in their preferences for explanation style, structure, and function. This suggests a complementary need for personalization, where explanation systems adapt to individual user profiles.
Future work should therefore explore both axes: generating a wider variety of faithful explanations, and tailoring them to the preferences and expectations of each user.

Beyond individual personalization, future systems could also adapt at the session or task level, selecting explanations dynamically based on context, cognitive load, or user engagement signals. We believe that LLMs, when coupled with inherently interpretable models, offer a unique opportunity to reach this level of adaptivity: they can generate varied narratives grounded in the same underlying mathematical factors, potentially addressing both the need for faithfulness and the demand for engaging content.

In sum, this work demonstrates that it is possible to build recommender systems that are interpretable by design and capable of generating user-friendly explanations. While our findings are promising, they also highlight the complexity of explanation design and the need for more adaptive, user-aware approaches. Bridging the gap between model transparency and perceived clarity remains a central challenge—one that future work must continue to address through interdisciplinary methods combining recommender systems, cognitive psychology, and natural language generation.

\begin{acknowledgments}
We thank all participants for their time and valuable feedback. According to the policies of our institution, this type of study did not require formal ethics board approval. Nevertheless, all participants were properly informed about the purpose of the study, their rights, and how their data would be used, in accordance with ethical guidelines.

The present research benefited from computational resources made available on Lucia, the Tier-1 supercomputer of the Walloon Region, infrastructure funded by the Walloon Region under the grant agreement n°1910247.
\end{acknowledgments}

NG acknowledges the support by the European Union (ERC consolidator, eLinoR, no 101085607).

\section*{Declaration on Generative AI}

During the preparation of this work, the authors used GPT-4o for the following activities: Grammar and spelling check and Paraphrase and reword. After using this tool, the authors reviewed and edited the content as needed and take full responsibility for the publication’s content.
In addition, large language models were used for experimental design purposes, specifically to generate explanations for recommendations, as detailed in the paper. This usage is mentioned here for transparency, and is distinct from the writing assistance described above.

\newpage

\bibliography{sample-ceur-cleaned} 

\appendix

\end{document}